# A general framework for the IT-based clustering methods

Teng Qiu     (qiutengcool@163.com)
Yongjie Li*      (liyj@uestc.edu.cn)

Key Laboratory of NeuroInformation, Ministry of Education of China, School of Life Science and Technology, University of Electronic Science and Technology of China, Chengdu, China
*Corresponding author.

**Abstract:** Previously, we proposed a physically inspired rule to organize the data points in a sparse yet effective structure, called the in-tree (IT) graph, which is able to capture a wide class of underlying cluster structures in the datasets, especially for the density-based datasets. Although there are some redundant edges or lines between clusters requiring to be removed by computer, this IT graph has a big advantage compared with the *k*-nearest-neighborhood (*k*-NN) or the minimal spanning tree (MST) graph, in that the redundant edges in the IT graph are much more distinguishable and thus can be easily determined by several methods previously proposed by us.

In this paper, we propose a general framework to re-construct the IT graph, based on an initial neighborhood graph, such as the k-NN or MST, etc, and the corresponding graph distances. For this general framework, our previous way of constructing the IT graph turns out to be a special case of it. This general framework 1) can make the IT graph capture a wider class of underlying cluster structures in the datasets, especially for the manifolds, and 2) should be more effective to cluster the sparse or graph-based datasets.

## 1 Introduction

### 1.1 Decision graph (DG)

In 2014, Alex and Alessandro (1) published a clustering method in science magazine titled "clustering by fast search and find of density peaks", which is theoretically simple yet sound, and technically fast, effective and reliable. Their method should belong to the class of density-based clustering methods, for which the cluster centers are usually associated with the density peak points.

It is non-trivial to detect the density peaks.

**Previously** (2-4), the density peak points are searched based on the "***zero-gradient***" **(ZG)** assumption, that is, the density peak points are of zero gradient in an assumed density function. A representative and popular approach is the one called Mean-Shift (2, 3). Based on the **ZG** assumption**,** Mean-Shift provides a simple iterative way to search the density peaks. Although Mean-Shift is effective, it confronts some well known problems: (i) applicable only to data points in Euclidean space; (ii) computationally costly and (iii) sensitive to the density ripples[1] or undesired local

---

[1] Namely, if the density function is not fitted well or with some density ripples due to the noise, each extreme point in ripples is also of zero gradient and thus will be falsely treated as cluster center. See also the last paragraph in Alex and Alessandro' s paper.

extreme points.

**In contrast**, the density peaks in Alex and Alessandro's method are searched based on a new assumption as follows:

> *compared with their neighbors (or the points in the same clusters), density peak points should have a relatively (i) high density and (ii) large distance from any points with higher density (usually in other clusters).*

Accordingly, each data point is featured by two variables: the density $\rho$ and the distance $\delta$, and only density peak points should have both high values in $\rho$ and $\delta$. Consequently, the density peak points pop out as outliers (e.g., the 1st and 10th points in Fig. 1B) in the 2-dimensional (2D) scatter plot coordinated by $\rho$ and $\delta$. This 2D scatter plot is called the "Decision Graph" (**DG**). Because of the visualization of the DG, those pop-out points (or outliers) can be easily yet reliably determined by computer, in virtue of users' interactive operation instead of a sophisticated outlier-detecting algorithm. In turn, after determining those pop-out points in the DG, the corresponding (Fig. 1A) density peak points (or cluster centers) are then determined. The effectiveness of the DG clustering method has been demonstrated by Alex and Alessandro on different tests.

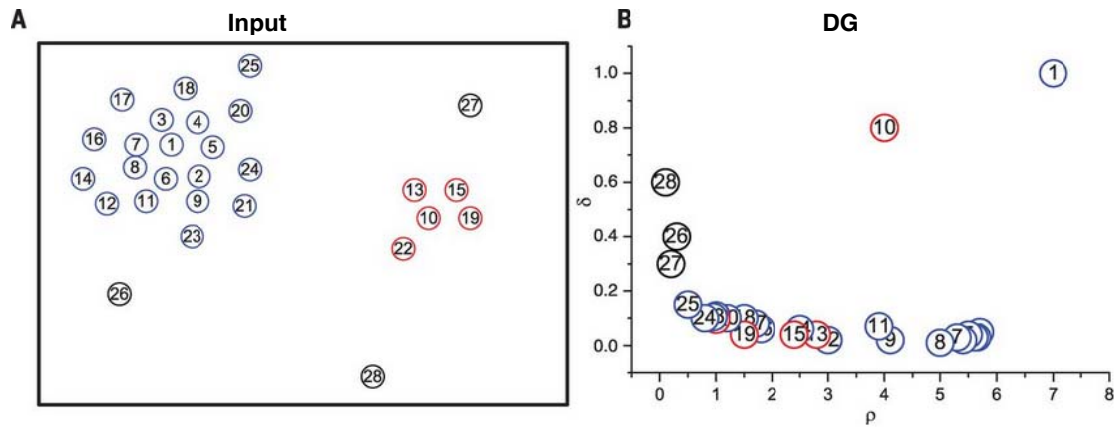

**Fig. 1. An illustration for the DG clustering** (1). (**A**) The input data points. Two clusters are denoted in red and blue, and the three outliers in black. (**B**) The DG. Two points from different clusters pop out and thus can be easily identified. In fact, those outliers (black), with a high value for $\delta$ and low value for $\rho$, can also be easily spotted or determined in DG.

Despite its success, DG clustering does not work well to a wider class of manifold datasets, especially for those arbitrarily twisted and uniformly distributed clusters, for which each cluster is hard to form one density peak point. This problem is also recently noticed in (5).

## 1.2 The in-tree (IT) based clustering methods

Also in 2014 and the early months of 2015, we proposed a serial IT-based clustering methods (6-9), which start with a physical imagination (6), in which the 2D space is treated as a horizontal rubber sheet and data points (or instances) as particles with mass. Intuitively, when a swarm of particles lie on the sheet, these particles will curve the sheet downward, and in turn the curved sheet will force particles to move from higher to lower potential areas. At last, these particles will cluster in certain places of locally lowest potentials. This process is actually a clustering behavior of a swarm of particles, and is applicable to the high-dimensional space and irrespective of the space being Euclidean space or not, thus motivating us to devise a similar approach to cluster data points.

The process of constructing the "curved" rubber sheet (or space) can be equivalent to the one of constructing the "non-uniform" space with varying "field" or "potential" based on the assumed rules like this: (i) each particle is assumed to generate a Gaussian field with its magnitude exponentially declined with the distance; (ii) the fields from different particles are additive. Our methods to mimic the particles' clustering behavior in this non-uniform space are different from that described in (10).

**Previously** (10), this kind of moving behavior is associated with the "force" concept, since Newton's force theory is so well-known. Then this "force" concept is mathematically modeled by "gradient", and thus data points stop or cluster at the places of zero gradient. Again, the **ZG** assumption appears. Similar to the case happens in previous density-based clustering methods we have mentioned in Section 1.1, this force-based method should also share the problems of Mean-shift clustering method.

**In contrast** (6), we don't seek the analytic solution or view the particles' clustering behavior analytically. Instead, we view it in a more general and abstract level. Consequently, "force" or "gradient" concepts are not involved. Instead, a simple yet general behavior rule is introduced, that is, particles have the tendency of moving from higher to lower potential areas, or the "descending tendency". Moreover, it is required that no particle can move from one place to another without a "process". Here, "process", in a continuous space, should be a smooth trajectory, whereas in a discrete data's space here, it can be approximated by a "zigzag" path, along which the data points serve as the transfer stations. How to choose the transfer station? the "proximity principle" is a natural choice. According to the "descending tendency" and the "proximity principle", the moving behavior for all data points at last becomes a very concise rule:

*each point "descends" to the nearest neighbor.*

We call it "the nearest neighbor descent" (**NND**) rule. 1) "descent" refers to the "descending tendency" from high to low potential areas, and 2) "the nearest neighbor" corresponds to the "proximity principle". As shown in Fig. 2B, if we treat the descending action in the NND rule by a directed edge, a fantasy graph structure appears. This fantasy structure proves to be an in-tree (**IT**) graph, as detailed in (6). Also see Section 3.1 for the definition of the IT graph.

Note that, in the NND rule, we treat all nodes equally in a general perspective. In other words, the particularities of the extreme points (with regard to the potentials) are temporarily ignored or left behind. That is why the local extreme problem that gradient-based methods confront is not involved here. Instead, it leaves the redundant edges (e.g., the purple one in Fig. 2B) problem. For clustering purpose, these redundant edges need to be determined or removed by computer and consequently (Fig. 2C) each independent sub-graph will represent a cluster. However, It is usually very easy for computer to determine the redundant edges, since they are usually more salient (or distinguishable) than the other edges.

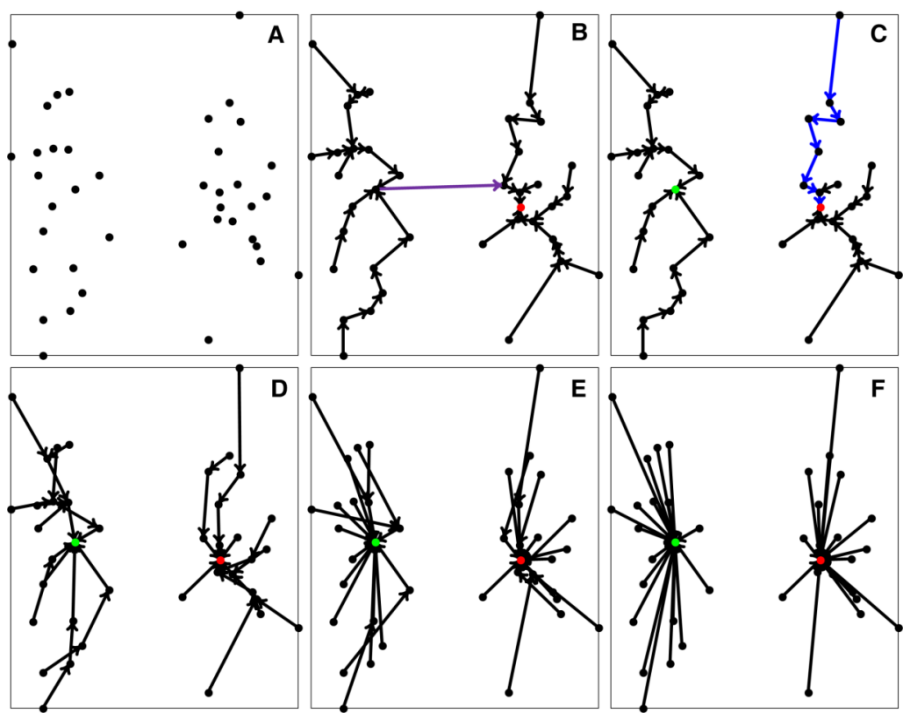

**Fig. 2. An illustration for our previous IT based clustering framework** (6). (**A**) The input data points. (**B**) The IT graph. The red point is the root node. For clustering purpose, the redundant or undesired edge (in purple) between clusters needs to be removed. (**C**) Two independent IT sub-graphs are obtained after removing the purple edge in (B), each being a cluster. Each node in any of the sub-graphs has one and only one path to reach the corresponding root node (red or green). (**D~F**) The process of finding the root nodes (or cluster centers) along the edge directions. The data points have same root nodes in (F) are assigned in same clusters.

In order to let readers deeply realize this "salient" feature of the redundant edges (or the saliency of the imperfect place) in the IT graph, we make a comparison between IT graph with three common neighborhood graphs: the minimal spanning tree (MST) (11), the Delaunay triangulation (DT) (12) and the k-nearest-neighbor

(k-NN) graph, as shown in Fig. 3.

In Fig. 3, we can see that, since the two elongated clusters (Fig. 3A) are very close and contaminated by noise, the redundant edges between clusters in the neighborhood graphs (Figs. 3B~D) are very short and thus it is hard for computer to distinguish them from the other edges. In contrast, the redundant edge between two clusters in the IT graph (Fig. 3E) looks much longer, since it doesn't connect the neighboring nodes, and instead, it usually starts from the node in the center of one cluster and ends to one node in another cluster, and we can also notice that the rest edges in the IT graph still connect neighboring nodes. This makes the redundant edge in Fig. 3E distinguishable from the other edges. Moreover, compared with the other neighborhood graphs, there are values (namely the potentials, denoted in different colors) assigned to the nodes in the IT graph. All these information helps us to devise easy yet effective rule to let the redundant edges in IT be determined and removed by computer.

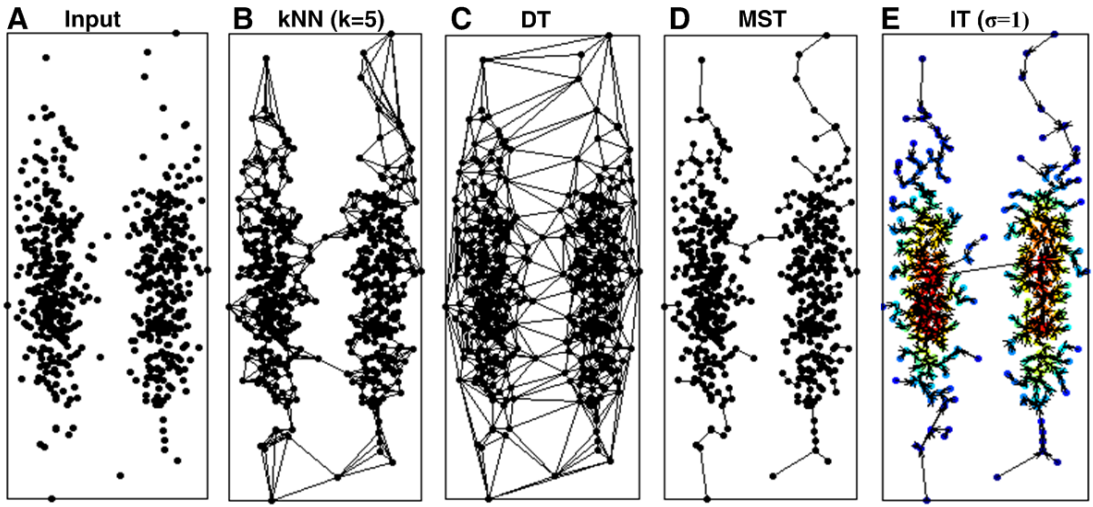

**Fig. 3. A comparison to show the saliency of the redundant edge in the physically-inspired IT graph.** The colors on nodes in (E) denote the magnitudes of the potential values. The redder the nodes, the larger the potential magnitudes are.

Moreover, if we ignore the redundant edges in Fig. 3E (or Fig. 2B), the clusters revealed in the IT graph are well in line with the underlying cluster structure in the data sets. Therefore, in conclusion,

> *1) This physically inspired IT graph well captures the underlying cluster structure in the data sets, except one imperfect place that there exist some redundant edges between clusters. And it is not hard for computer to determine those redundant edges, due to the saliency feature of them.*
>
> *2) The whole clustering process follows the methodology of from generality (referring the NND rule) to particularity (referring the edge-cutting issue).*

In fact, we have made some attempts in the edge-cutting issue. In (6), the interactive (Int-Cut), and semi-supervised (Sup-Cut) methods were proposed to cut those redundant edges. In (7), we combined our IT structure with affinity propagation (AP) (13) and consequently an automatic cutting method, called **G-AP**, was proposed therein. Interestingly, G-AP turns out to be more powerful than AP, since G-AP can discover the non-spherical clusters that AP cannot. In (8), we combined our IT structure with the isometric feature mapping (Isomap) (14) and consequently an interactive method, called **IT-map**, was proposed therein. Also interestingly, IT-map can preserve the clusters while mapping data points into the low-dimensional Euclidean space, whereas this is hard for Isomap due to the so-called crowding problem. Moreover, in our first paper (6) of this serials, we also demonstrated in details that Alex and Alessandro' s Decision Graph method can actually be viewed as another very simple interactive and reliable method to remove those redundant edges. Because, simply by using two intermediate variables in IT graph, i.e., the potential and edge distance, we can also derive a similar DG, in which the pop-out points actually correspond to the start nodes of those redundant edges. All these methods above constitute a group of the **IT clustering methods or IT-based clustering methods**.

In fact, we have to admit that it's an amazing coincidence that our NND rule turns out to be very similar to that of Alex and Alessandro's DG clustering method. That is why that, on one hand, their method can be one of the methods to determine the redundant edges in our IT graph; on the other hand, our methods can be viewed as a complement of theirs, in both theory and methodology levels: in theory, the physical background and graph-based implementation in our first paper provides the physical and graphical explanation for the efficiency behind the DG clustering method, and thus increasing our understanding to DG; in methodology, compared with DG, our "graph" perspective provides a concrete problem, i.e., the redundant edge problem, and consequently we are motivated to extend a "tree" (referring the DG clustering method or our first paper) to a "forest" (referring the group of IT-based clustering methods). In other words, **this concrete "problem" can lead to more "answers"**.

In this work, we will continue to extend this IT-based clustering methods family. However, rather than propose new methods to remove those redundant edges, we will propose a general way to construct the IT graph.

## 2 Motivation

We notice in our first paper that, we actually defaults to consider all nodes in a "complete graph", which means that each node is connected directly with all the other nodes by edges[2], and thus this complete graph actually ignores the underlying structure information in datasets. We also notice in Fig. 3 that, although, like the IT graph, if we ignore those redundant or undesired edges between clusters, those neighborhood graphs (i.e., the *k*-NN, MST, DT) can also capture the underlying structure (two clusters) in the data set in a sparse form. And in fact, those neighborhood graphs are more capable of doing that, since they rely only on the pair-wise distance information to construct the graphs, and thus being of general meaning to a wider class of datasets. This is reminiscent of the famous manifold method—Isomap (14), which builds the classic multidimensional scaling (MDS) dimensionality reduction method on the k-NN graph, and consequently Isomap is able to deal with an important class of data manifolds.

---

[2] However, the similarity can be 0. In other word, our previous framework can also deal with the sparse distance or similarity matrix.

Here, we will also build our previous framework on the neighborhood graphs (e.g., Fig. 3B~D) instead of the complete graph, so as to make all the IT-based clustering methods be able to deal with a larger range of clustering problems.

## 3 Method

### 3.1 Preliminaries

"**Graphs**", in Graph theory, are the structures in which each data point (or instance) is viewed as a node, and the relationship between each pair of data points is represented by an edge. The edge length (or weight) usually represents the distance (or similarity) between the data points. Note that the "graphs" here are just imaginary or auxiliary models to help us understand the principles of the methods. The following IT graph and the neighborhood graphs belong to this scope.

**The in-tree (IT) graph**, also called in-arborescence or in-branching graph (15, 16), is a directed graph that meets the following conditions: (i) only one node (also called the root node) has no directed edge started from it (also called outdegree 0); (ii) any other node has and only has one directed edge started from it; (iii) there is no cycle in it; (iv) it is a connected graph. In brief, the IT graph is a connected, directed, and acyclic graph, together with beautiful order (namely, for the root node, each other node has one and only one directed path to reach it).

**The neighborhood graphs** can be either parametric or non-parametric. Typical parametric neighborhood graphs are the *k*-nearest-neighborhood (*k*-NN) and the $\epsilon$-nearest-neighborhood ($\epsilon$-NN) graph. For *k*-NN graph, each node *i* selects as its neighbors the *k* nearest nodes. For $\epsilon$-NN graph, each node *i* takes as its neighbors any node *j* within the radius $\epsilon$, i.e., $d_X(i,j) < \epsilon$. Typical non-parametric neighborhood graphs are the MST and DT as mentioned in Fig. 3, together with the relative neighborhood graph (RNG) (17), Gabriel graph (GG) (18). These non-parametric graphs are all the connected graphs. They are all designed to find the underlying structure in data set, only relying on the pair-wise distances or similarities while with different rules or goals. For instance, the MST is a graph that aims to connect all nodes with the least sum of the edge lengths. Interestingly, they have such relationships $MST \subseteq RNG \subseteq GG \subseteq DT$ (19), namely, the MST is the sub-graph[3] of the RNG, and the RNG is the sub-graph of the GG, etc.

---

[3] By removing some edges in RNG, one can get MST.

Note that, the **Decision Graph** (DG) in Section 1.1 refers the 2D scatter plot, different from the "graph" terms in graph theory.

**3.2 The proposed method in details**

The proposed method takes as input the distance $d_X(i,j)$ of each pair of data points $i$ and $j$ $(i, j = 1, 2, \cdots, N)$, and contains six steps detailed as follows (***see also an overview in Table 1 and an illustration in Fig. 4***):

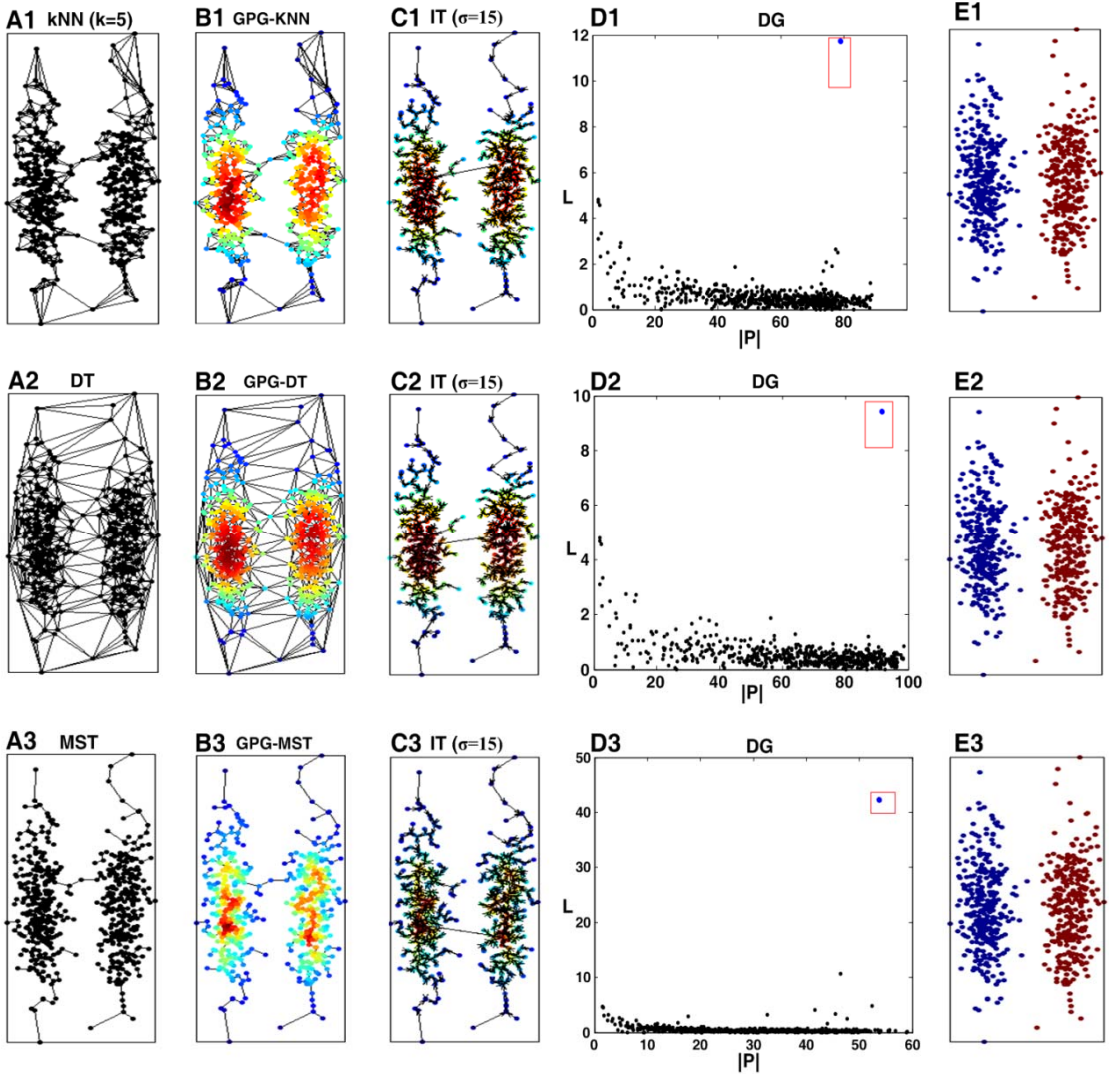

**Fig. 4. An illustration for the proposed method.** (**A1~A3**) The *k*-NN graph (*k* = 5), DT and MST, respectively. The corresponding global potential graphs (**GPG**) are shown in (**B1~B3**), and the corresponding IT graphs are shown in (**C1~C3**). Colors on nodes in (B1~B3) and (C1~C3) denote the magnitudes of potentials. (**D1~D3**) The Decision Graphs (**DG**) corresponding to the IT graphs in (C1~C3), respectively. The red boxes are drawn by the user, and computer will judge the points (blue) inside them as the pop-out points. These pop-out points correspond to the start nodes of the those redundant edges in (C1~C3). (**E1~E3**) Clustering results. Data points in same colors are assigned to the same clusters.

**In the 1st step**, the neighborhood graph, denoted as G, is constructed, There are many choices for the neighborhood graph, e.g., the k-NN, ϵ-NN, or the MST, RNG, GG, DT, etc.

**In the 2nd step,** the graph distance $d_G(i,j)$ is computed for each pair of nodes $i$ and $j$ in the graph G. Here, the graph distance refers the shortest path distance. Therefore, if G is not a connected graph and nodes $i$ and $j$ are in different sub-graphs, then $d_G(i,j) = +\infty$, since there is no path between them. This graph distance can be computed by the classic Floyd (20) or Dijkstra (21) algorithm, and a fast approach is provided in (22) which makes good use of the sparseness of the Graph.

**In the 3rd step**, the potential $P_i$ is computed by

$$P_i = -\sum_j \exp(\frac{-d_G^2(i,j)}{\sigma}); \quad j=1,2,\cdots,N; j \neq i \tag{1}$$

Where $\sigma$ is a parameter that can be adjusted by the users. Note that, compared with our previous work (6), there are two differences here: (i) the graph distance $d_G(i,j)$ instead of the input distance $d_X(i,j)$ is used; (ii) the distance is squared (this can obtain better performance).

Here, we call the graphs with graph-distance-based potential values on nodes as the global potential graphs (GPG). Based on different initial graphs as *k*-NN, MST, and DT (Fig.3B~D), different global geometric potential graphs as **GPG**-kNN, **GPG**-MST, **GPG**-DT, can be obtained, as shown in Fig. 4 B1~B3, respectively, where the potential values are denoted by different colors.

**In the 4th step**, each node *i* "descends" to the nearest node. This unique node is defined as

$$I_i = \arg\min_{j \in J_i} d_G(i,j) \tag{2}$$

Where $J_i = \{j \mid P_j < P_i\}$. That is, $I_i$ is the nearest node (regarding graph distance) among those with lower potential values ("descending direction") compared with node *i*.

We define $I_i = i$, if node *i* has the lowest potential. It is suggested to add the data index term in $J_i$ as what we did in (6) where $J_i = \{j \mid P_j < P_i\} \cup \{j \mid P_j = P_i \,\&\&\, j < i\}$, which will bring some advantages. Moreover, this step is actually an approximation to the traditional gradient-based methods and consequently the local extreme problem of the gradient-based methods is largely reduced.

If we connect each node *i* to node $I_i$ by an directed edge (or line), the IT graph[4] is constructed, as shown in Fig. 4 C1~C3, which contains *N* nodes and *N* - 1 directed edges (note that there is one node for which $I_i = i$, it points to itself and this edge is ignored here). For each directed edge, denoted as $e(i, I_i)$, node $i$ is the start node and node $I_i$ is the end node. The length of the edge $e(i, I_i)$ is denoted as $L_i = d_G(i, I_i)$ here.

---

[4] Note that, if the initial graph G is not connected, mainly for k-NN or ϵ-NN graph, there will be several IT sub-graphs.

In the 5th step, the redundant edges in the IT graph need to be determined or removed by computer. Several methods can be used, e.g., Alex and Alessandro's Decision Graph (DG), or the methods proposed in our previous works as IT-maps (8), G-AP (7), Int-Cut (6), Sup-cut (6). Here, we use DG to (i) demonstrate the effectiveness of this work and (ii) to show the connection of DG with our framework. We mainly use two features, $P_i$ and $L_i$, in IT graph associated with any node $i$. $P_i$ denotes the potential of node $i$ (see step 3) and $L_i$ is the length of the directed edge started from node $i$ (see step 4). Thus, we can map each node to a similar Decision Graph coordinated by those two variables, as shown in Fig. 4 D1~D3 (note that we use the magnitude for $P_i$). From Fig. 4 C1~C3, we can see that only the start nodes of the undesired edges (between clusters), denoted as $e_u(i, I_i)$, has high values in both $L_i$ and $|P_i|$ and thus will pop out in Fig. C1~C3. In turn, from the DGs, those undesired or redundant edges can be indirectly determined by their start nodes, since each node in the IT graph is the start node of only one directed edge[5].

DG is an effective interactive method to help computer determine those redundant edges in the IT graph, since we can always "see" the start nodes of those redundant edges in the 2D scatter plot, irrespective of whether the IT graph or the redundant edges can be visualized or not.

After removing those undesired edges by computer, each sub-graph is in fact a cluster, and thus those nodes belonging to same sub-graphs are assigned to same clusters.

**In the last step**, computer will determine the members (i.e., the nodes) in each sub-graph. Since each sub-graph is still an IT graph, each with a root node (can be viewed as the cluster center), so computer can first let all nodes find their corresponding root nodes. Since each node in IT graph has one and only one directed path to reach the root node, so by searching along the directed edges, each node can be sure to find their roots in finite steps. To be specific, in step 4, we know that node $I_i$ is the first transfer node that each node $i$ "descends" to, then where is the next transfer point for each node? It is where the node $I_i$ "descends" to, that is, $I_{I_i}$. In other words, the next transfer nodes for all node are $I_{I_i}, i=1,2,\cdots N$, or $I_I$. So, $I$ here has two functions: one is to store the end node of each node $i$; the other is to serve as an index for the next transfer nodes. This index $I$ is updated each time by $I_I$ and is sure to stop in very few steps. As proved in (6), $\lceil \log_2(H) \rceil$ steps are need, where $H$ is the maximum number of edges of all directed paths in the graph after cutting. The stop criterion is that no change happens for the updating. At last, $I_i$ denotes the root node of the sub-graph where node $i$ resides.

---

[5] Note that, for IT graph, each node is just the start node of only one directed edge, but each node is usually not the end of only one directed edges. Therefore, the directed edge can be indirectly determined by its start node.

**In conclusion** (see also a brief expression in Table 1), the 1st and 2nd steps are similar to those in Isomap, except that there are more choices (*k*-NN, MST, DT, etc.) in constructing the initial graph in step 1. These two steps serve to obtain the graph distance $d_G(i,j)$. The 3rd and 4th steps are almost the same as what we did in our first work except that the graph distance $d_G(i,j)$ instead of the input $d_X(i,j)$ participates in the computation here. The last two steps are the same as our previous works[6], except the fact that we now have more methods that can be used to remove the redundant edges in step 5.

**Table 1. An overview of the general framework for the IT-based clustering methods**

**Input:** Distance $d_X(i,j)$.
**Output:** $I$.   // vector $I$ stores the roots for all nodes.
**Steps:**
1, **Construct the neighborhood graph** $G$    // $G$ can be *k*-NN, MST, DT, GG, RNG, etc.
2, **Compute the graph distance** $d_G(i,j)$    // the shortest path distance
3, **Compute the potential** $P_i$

$$P_i = -\sum_j \exp(\frac{-d_G^2(i,j)}{\sigma}); \quad j = 1,2,\cdots,N; j \neq i.$$

4, **Construct the IT graph**    // "descend to nearest neighbor"
   Define the end node of each directed edge $e(i, I_i)$:

$$I_i = \arg\min_{j \in J_i} d_G(i,j), \ i = 1, \cdots, N$$

   where $J_i = \{j \mid P_j < P_i\}$. And $I_i = i$, if $P_i$ is the lowest.
   The edge length for each directed edge $e(i, I_i)$ is defined as $L_i = d_G(i, I_i)$
5, **Remove the redundant directed edges** $e_u(i, I_i)$
   determine $e_u$ or its start node *i* by **Methods**: DG, IT-maps, G-AP, Int-Cut, SS-Cut, etc**.**
   replace the end node $I_i$ of $e_u$ by its start node *i*.    // equivalent to removing $e_u$
6, **Find the roots**
   Update $I$ by $I_I$ until $I == I_I$. // "Search along the directed edges, until reach the roots"

---

*Words behind "**//**" are some annotations
*Abbr. **KNN:** k-nearest neighborhood;   **MST:** minimal spanning tree;   **DT:** delaunay triangulation;
       **RNG:** relative neighborhood graph (RNG); **GG: Gabriel graph**
       **IT**: in-tree;   **DG:** decision graph;   **G-AP:** generalized affinity propagation;
       **Int-Cut:** interactive cutting;   **SS-cut:** semi-supervised cutting.

## 4 Experiments

We first tested the two-Gaussian dataset appeared in Fig. 3A, using three different neighborhood graphs, the k-NN, MST, and DT, respectively. In Fig. 3B~D, we have seen that the redundant edges for these neighborhood graphs are very non-salient,

---

[6] Therefore, the last step in the new framework can refer to Fig. 2 D~F for illustration.

whereas the proposed method can transfer them to the salient IT graphs as shown in Fig. 4 C1~C3, respectively. In fact, these results can be robust to a wide range of choices on the parameter σ, as shown in Table 2. For instance, when the MST was used in step 1, we arbitrarily chose 9 different values among the interval (1, 200) for σ, all leading to the excellent results, i.e., the clustering results are consistent with visual perception, and the IT and DG graphs are also very salient. "salient" means the redundant edges or the pop-out points are very distinguishable, as those results in Fig. 4. Note that, The DG graphs here, as shown in Fig. 4 D1~D3, are little different from those in Alex and Alessandro's paper (as in Fig. 1B). Here, the number of pop-out points is one less than the number of clusters obtained.

Table 2. Tests on two Gaussian datasets

| Neighborhood graph | k | σ | performance |
|---|---|---|---|
| k-NN | 5 | 1, 5, 15, 20, 25, 40, 80, 100 | excellent* |
| MST | — | 1, 5, 10, 20, 40, 60, 80, 100, 200 | excellent |
| DT | — | 1, 5, 10, 20, 40, 60 | excellent |

*excellent (the same in other Tables) and the clustering results are consistent with visual perception, and the DGs are also very salient.

We also tested different shapes of datasets, as shown in Table 3, where the initial graphs are all the k-NN graphs, with fixed values for *k* while varying values for *σ*. We also arbitrarily chose several values for *σ* in as large range as possible. For instance, for the spiral dataset, *σ* was set as small as 1, and as large as 100. Although, for the Jain data set, the range for the appropriate σ is relatively narrow, this is still a progress, since this dataset is not easy to be clustered either in our previous framework or Alex and Alessandro's Decision Graph. Figure 5 lists one test result for each data set.

Table 3. Tests on different datasets when k-NN is used

| Data sets | k | σ | performance |
|---|---|---|---|
| Spiral (23) | 5 | 1, 2, 5, 10, 100 | excellent |
| Flame (24) | 5 | 2, 3, 5, 10 | excellent |
| Aggragation (25) | 30 | 5, 8, 10 | excellent |
| Jain (26) | 10 | 40, 50 | Good* |
| Compound (11) | 10 | 20, 25 | Good |

*Good (the same in other Tables) means the clustering results are consistent with visual perception, whereas the DGs are not very salient.

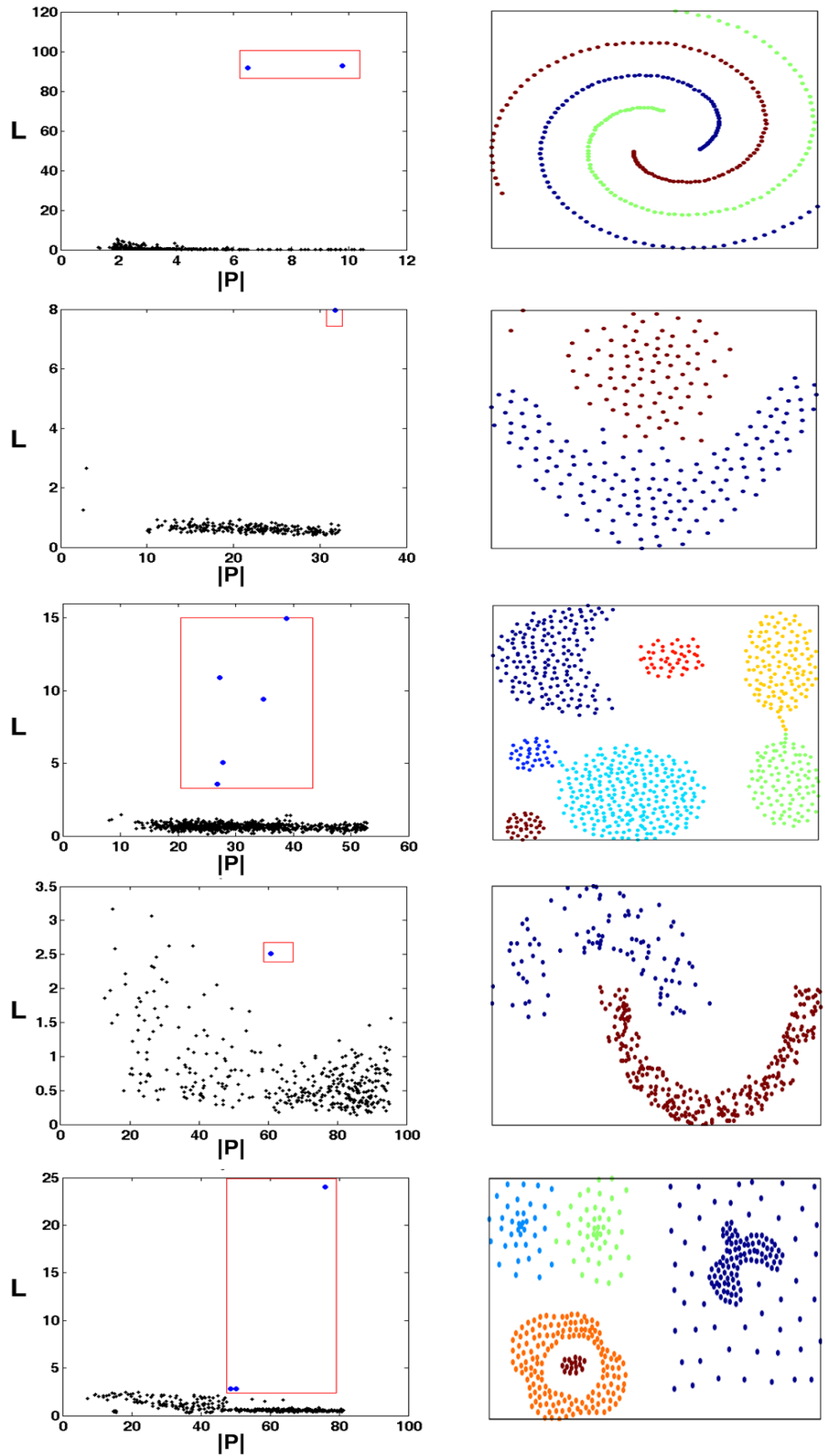

**Fig. 5. Several results on different datasets.** Each row shows the Decision Graph (left) and clustering result (right) of each data set. **From up to bottom**, data sets: the Spiral, Flame, Aggragation, Jain and Compound datasets; parameters (k, σ): *k* = 5, 5, 30, 10, 10; *σ* = 1, 5, 8, 50, 25.

We also tested one dataset from Table 3, with varying values for *k* and fixed value for *σ*. Here, *k* refers to the parameter in k-NN graph. As shown in Table 4, the results are quite robust to the choice for *k*. See details for the results when *k* = 8, 55, 200 in Fig. 6. Note that in Fig. 6 A1, the k-NN graph is not connected, with five independent sub-graphs. Therefore, there are only two pop-out points in Fig. 6 C1, corresponding to the two undesired edges in two sub-graphs in Fig. 6 A1.

Table 4. Tests on Aggragation data with different k

| Data set | k | σ | performance |
|---|---|---|---|
| Aggregation | 8, 35, 55, 100, 200 | 7 | excellent |

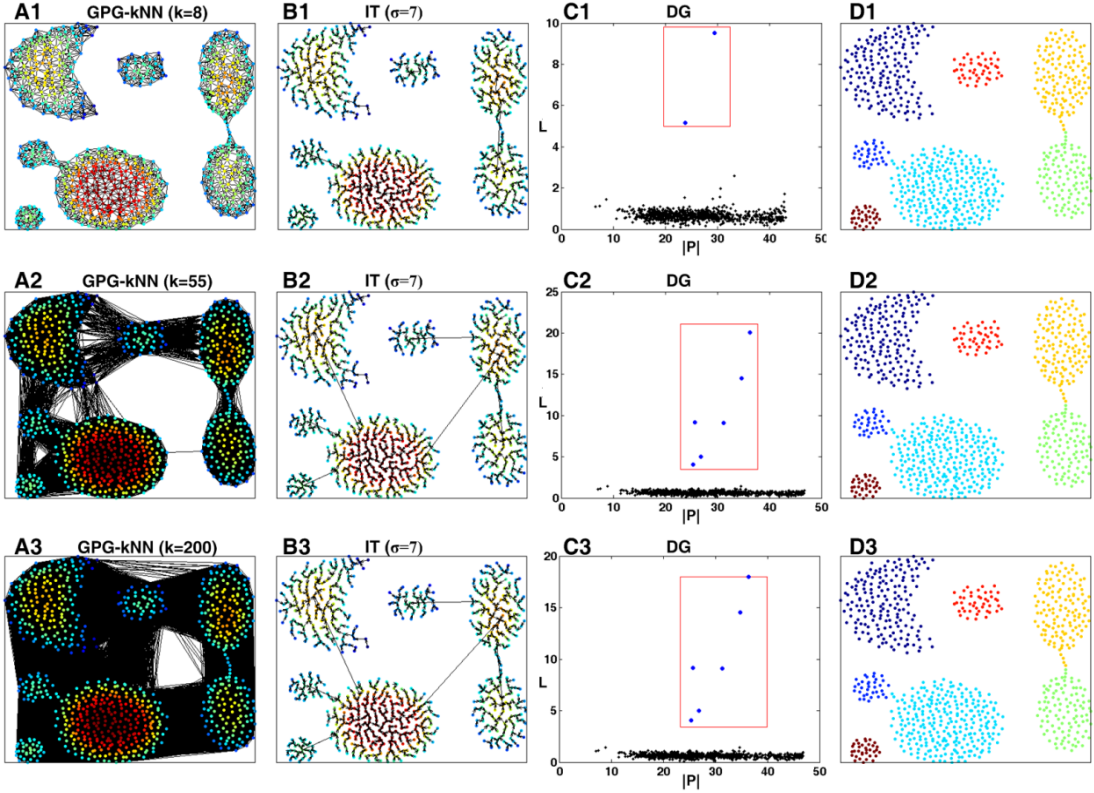

**Fig. 6. The experiments with varying *k* and fixed *σ* =7.** The results when *k* = 8, 55, 200 are shown in the 1st (A1~D1), 2nd (A2~D2) and 3rd (A3~D3) rows, respectively.

## 5 Conclusions and Discussions

The proposed general framework makes an effective combination of the neighborhood graphs and the NND rule by taking the graph distance $d_G(i,k)$ (in steps 2~4 in table 1) as a bridge. Consequently, it inherits the advantages of both the neighborhood graphs and the NND rule, i.e., (i) the neighborhood graphs are flexible to capture the underlying clustering structures in the datasets; (ii) the NND rule can construct the IT graph whose redundant edges are easy to be determined. In effect, as in Fig. 4, the underlying structure is first captured in one of the neighborhood graphs (e.g., the KNN, MST, DT graph in Fig. 4 A1~A3), which is then transformed by the NND rule to a salient graph (i.e., the IT graphs in Fig. 4 C1~C3). "Salient" refers that the redundant edges are very easy to be determined. In fact, we can see in

Fig. 4 D1~D3 that, the edge length *L* in the IT graphs can serve as a suitable measure to make the redundant edges salient enough to be distinguished from other edges. This is a progress in structure, since, for the MST, KNN, or DT, we usually cannot rely on the edge length to determine those redundant edges.

Compared with previous framework, this new geometric framework (Table 1) has at least the following advantages:

(i) the new framework can make the IT graph capture a wider class of underlying clustering structures in the datasets. Consequently, as shown in Fig. 5, the DG method (viewed as one edge-removing method in step 5) is now able to detect the clustering structure in Jain and Compound datasets,.

(ii) The new framework becomes more flexible and powerful. As shown in Table. 1, there are not only multiple choices to remove the redundant edges (step 5), but also multiple choices for constructing the initial graph (step 1). Certain neighborhood graph should be superior to other graphs in dealing with some specific problems. Note that when the MST graph is used in step 1, the corresponding graph distance is also called the Minimum Curvilinearity distance (27, 28).

(iii) According to the experiments and the principle (the graph in step 1 approximately captures the underlying structure of data sets), this new framework should be more robust to the parameter σ. We will provide more evidence for this in the future.

(iv) We believe the proposed general framework can reveal more meaningful underlying structures or relationships in the datasets when the input distance (or similarity) matrixs are sparse or when the datasets are graph-based. Because, previously, the distance between the points *i* and *j* without direct connection will be considered as $+\infty$. In contrast, in this new framework, since points *i* and *j* may still be indirectly connected by a "path" in the graph, and thus the distance between them **will not** and **should not** be $+\infty$.

In fact, the previous framework is a special case of the proposed one here, that is when *k* = *N*-1, the case that the graph distance $d_G(i,k)$ (in steps 2~4 in table 1) reduces to the pair-wise distance $d_X(i,k)$ between data points.

Here, we mainly consider this general proposal in our graph perspective. Of course, this can also be illustrated from Alex and Alessandro's perspective by defining the two variables for each node *i* like this: the local density $\rho_i = \sum_j \exp(-d_G^2(i,j)/\sigma)$ and the distance $\delta_i = \min_{j:\rho_j > \rho_i} d_G(i,j)$.